\documentclass[letterpaper, 10 pt, conference]{ieeeconf}  
\IEEEoverridecommandlockouts                              
\usepackage{amsmath}
\usepackage{tabularx}
\usepackage{url,lineno,microtype}
\usepackage{float}
\usepackage{placeins}
\let\labelindent\relax
\usepackage[inline]{enumitem}
\usepackage[font={small}]{caption}
\usepackage{siunitx}            
\usepackage{graphicx}
\usepackage{subcaption}
\usepackage{xcolor}
\usepackage[bibstyle=ieee,citestyle=numeric-comp,giveninits=true,doi=false,maxbibnames=3]{biblatex}
\usepackage[official]{eurosym}
\DeclareGraphicsExtensions{.png,.eps,.pdf,.jpeg,.jpg}

\title{\LARGE \bf FootTile: a Rugged Foot Sensor for \\Force and Center of Pressure Sensing in Soft Terrain}
 
\author{Felix Ruppert and Alexander Badri-Spr\"owitz\thanks{Authors are with the Dynamic Locomotion Group, Max Planck Institute for Intelligent Systems, Stuttgart, Germany}\thanks{Correspondance: {\tt\small ruppert@is.mpg.de}}\thanks{FR contributed to concept, design, experiments, data analysis and writing. ABS contributed to concept and writing.}\thanks{We thank the International Max Planck Research School for Intelligent Systems (IMPRS-IS) for supporting the academic development of Felix Ruppert. This work was made possible thanks to a Max Planck Group Leader grant awarded to ABS by the Max Planck Society.}}

\begin{document}
\maketitle
\thispagestyle{plain} \pagestyle{plain}

\begin{abstract}
In this paper we present FootTile, a foot sensor for reaction force and center of pressure sensing in challenging terrain. We compare our sensor design to standard biomechanical devices, force plates and pressure plates. We show that FootTile can accurately estimate force and pressure distribution during legged locomotion. FootTile weighs \SI{0.9}{g}, has a sampling rate of \SI{330}{Hz}, a footprint of 10\,x\,\SI{10}{mm} and can easily be adapted in sensor range to the required load case. In three experiments we validate: first the performance of the individual sensor, second an array of FootTiles for center of pressure sensing and third the ground reaction force estimation during locomotion in granular substrate. We then go on to show the accurate sensing capabilities of the waterproof sensor in liquid mud, as a showcase for real world rough terrain use.
\end{abstract}

\section{INTRODUCTION}
In walking robotics and in biomechanics, sensors are a fundamental tool to understand the mechanics and control of a system. Biomechanic data is especially important in outdoor environments, to collect data during natural locomotion.
Standard tools in biomechanics, are ground reaction force plates and pressure plates to measure the force and pressure distribution on a foot. Force plates and pressure plates are easy to use, but are are heavy, delicate, immobile and expensive. It is also not possible to measure pressure distribution on soft and granular substrate. The biggest drawback of these systems, however, is that they are not wearable. Both systems can only capture a small number of strides because of their size. Therefore it becomes hard to collect average data or data where the conditions change between steps.\\
To have a pressure or force reading for a long duration at every step of a system, wearable devices are required. The most direct sensors are strain gauge based force/torque sensors \cite{Kaslin2018}. Attached to the foot, these sensors can read forces and torques with high range, accuracy, and frequency. Stiff sensors, however, tend to be heavy and increase the rotational inertia when placed at the end of a leg. They are also expensive and very sensitive to impact forces. Soft sensors solve the problems of weight and impact sensitivity by implementing soft, deformable elements. Under load, the soft element deforms, and various sensing principles can measure the deformation. The Optoforce  sensor measures the changes in light reflection on the inside of a soft dome due to deformation.
Alternatively, the deformation of a soft sensor body can be detected by Hall effect sensors  \cite{Ananthanarayanan2012a}, resistive \cite{Hutter2012, Takahashi2005, Fondahl2012, Lungarella2004} or capacitive \cite{Wu2016} sensors. While these sensors are easy to use and cheap to implement they are not sufficiently robust against the high impact forces during running and can not easily be scaled down and miniaturized.\\
\begin{figure}[t]
	\centering
  \includegraphics[width=\columnwidth]{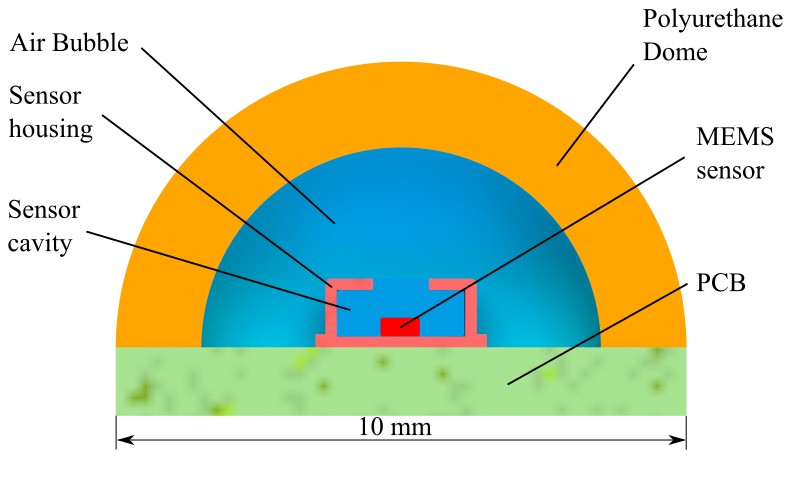}
	\caption{FootTile sensor design, inspired from \cite{Leon2019}, with significant changes to bubble shape, size, and simplified dome morphology. The size of the air bubble, and the shape of the dome influence the sensor's characteristics. The blue areas show the air bubble volume. External forces will compress the air, and increase the air pressure. The FootTile in this image weighs \SI{0.8}{g}, a size of 10 x 10 x \SI{5.5}{mm}, a sample rate of \SI{330}{Hz} and a resolution of \SI{10}{bit} in \SI{30}{N}.}
	\label{fig:sensorDesign}
	\vspace{-4mm}
\end{figure}Tenzer and colleagues \cite{Tenzer2014} developed a tactile sensor called 'Takktile', based on a barometric pressure sensor. The micro-electro-mechanical (MEMS) sensor's cavity gets filled with rubber that transmits external forces to the pressure sensor's membrane. The miniature sensor can measure forces up to \SI{5}{N} and can be overloaded with \SI{400}{\%} of its nominal force. To extend the sensor range, Chuah and colleagues \cite{Chuah2014} implemented a higher range pressure sensor in the same design to circumvent the range limits. L\`eon and colleagues \cite{Leon2019} enhanced the sensor range by including an engineered bubble into the rubber dome. This bubble acts as a pressure transducer. Based on the dome deformation the bubble volume reduces, and the pressure in the bubble rises. The range can be designed based on the dome design and can be adjusted for any load case. The sensor was used in the paper to measure the forces on a rotorcraft robotic landing gear.\\
In this project, we adapt the sensor design from \cite{Leon2019}, for use in legged locomotion, with our requirements for force range,  sensitivity, sensor size and sampling frequency. Specifically, we implement a sensor array and showcase measurements of forces as well as the center of pressure on solid and soft terrain. We present a foot sensor array that is \emph{lightweight}, \emph{rugged}, \emph{wearable}, \emph{waterproof} and allows \emph{force and pressure sensing} in flat and rough terrain and granular media as well as mud. As shown in previous research, the sensor can be used on robotic systems for control purposes. We illustrate the advantages of this miniature sensor for use in biomechanical legged locomotion experiments for animal or robotic subjects. We compare the sensor performance to force plates and pressure plates, the two standard devices for biomechanical data acquisition. We show how the sensor can estimate ground reaction forces and the center of pressure. We present experimental data how the sensor can be used on solid ground, as well as granular medium and real world rough terrain, like mud.\\
\section{METHODS}
\subsection{Sensor design}
The sensor in \autoref{fig:sensorDesign} consists of a barometric pressure sensor (MPL115A2, \textit{NXP}) . To date, it is the miniature barometric pressure sensor with the highest sample rate on the market. 
In this paper we implement simplified polyurethane half-spheres with a spherical bubble inside to reduce the mold complexity. By using different PU materials, three design parameters can be used to adapt the sensor range to the application. The spheres are produced in a two-part 3D printed mold (Onyx, \textit{Markforged}) and glued airtight to the sensor PCB with instant glue. To make the FootTile as small as possible, we chose a bubble diameter of \SI{6}{mm} to fit around the sensor diagonal of \SI{5.4}{mm}. The total sensor design has production costs of less than 10 \euro{} per sensor.\\
The pressure sensor is internally temperature compensated with an individual calibration. Using temperature compensated pressure for calculations we can assume an isotherm process ($T=\textrm{const}$), the general gas law results in:
\begin{eqnarray}
p_1\cdot V_1 &=& p_2\cdot V_2 \big|_{T=\textrm{compensated}} \nonumber \\
\text{which leads to} \nonumber \\
p_2 &=& p_1\cdot\frac{ V_1}{V_2}
\end{eqnarray}
where $p_i$ is the pressure in the bubble, $V_i$ is the bubble volume, and $T$ is the gas temperature in the bubble, before and after deformation, respectively. The MEMS sensor has a range of \SI{50}{kPa} to \SI{115}{kPa}. The initial pressure in the bubble during manufacturing is one atmospheric pressure, around $p_1=$\SI{1}{bar}$\equiv$ \SI{100}{kPa}. The sensor saturates at $1.15\cdot p_1 = $\SI{115}{kPa}. To maximize the range of the MEMS sensor, the bubble should compress by $13$\% of its initial volume at the maximum external load. By changing the bubble volume and the dome geometry, the sensor range can be designed for many load cases.\\
To show the range dependency on the dome and bubble volume as well as the material, we test the parameter space shown (\autoref{tab:parameters}).
\begin{table}
\renewcommand{\arraystretch}{1.5}
\caption{Experimental parameters for sensor design}
\label{tab:parameters}
\begin{tabularx}{\columnwidth}{X X}
\hline
Parameter & Values\\
\hline
Dome diameter $d_{dome}$ [mm] & 10, 11, 12\\
Bubble radius $r_{bubble}$ [mm] & 3\\
Material $(E_{Young})$& Vytaflex40 (E=\SI{0.69}{MPa}), \\& Vytaflex60 (E=\SI{2.17}{MPa})\\
\end{tabularx}
\end{table}

\subsection{Sensor validation}
\begin{figure}
	\centering
  \includegraphics[width=\columnwidth]{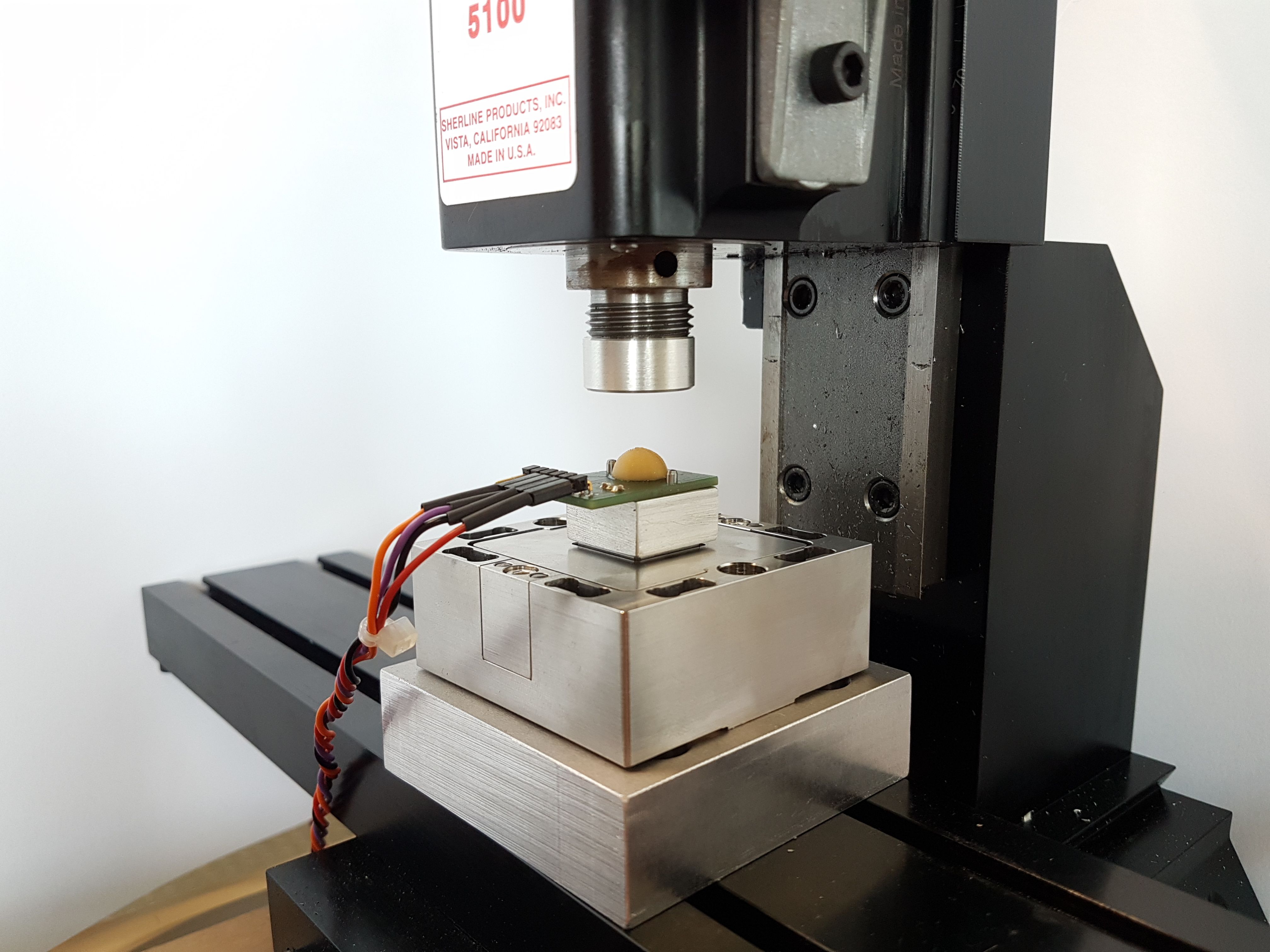}
	\caption{FootTile mounted to a force sensor for calibration. The guide of the manual mill guarantees purely vertical deflection through the indenter, fixed to the milling head. The FootTile is compressed until its barometric pressure sensor saturates.}
	\label{fig:sensorMill}
	\vspace{-4mm}
\end{figure}To calibrate the sensor reading to the applied force, we use a 3-axis force sensor (K3D60, \textit{ME-Systeme}) as ground truth. The FootTile is fixed to the force sensor mounted on the z-axis of a manual mill (\autoref{fig:sensorMill}). By moving the mill in the vertical direction, the indenter on the mill head deforms the FootTile, and the force sensor data can be compared to the FootTile data. The FootTile is connected via I$^2$C to a Raspberry Pi 3B+. The force sensor is connected to an amplifier (BA9236, \textit{Burster}) and an analog-digital converter (MCP3208, \textit{Microchip}). The different sensor configurations from \autoref{tab:parameters} are indented until the sensor saturates at \SI{115}{kPa} to determine the maximum sensor range. The sensor has a conversion time for one data sample of \SI{1.6}{ms}. This results in a maximum theoretical reading frequency of \SI{625}{Hz}. Due to speed restrictions in the Raspberry I$^2$C driver, data for both sensors is sampled at \SI{330}{Hz}. The indentation experiments are repeated four times, and the data is averaged and shown with 95\% confidence intervals.\\

\subsection{FootTile array}
\begin{figure}
	\centering
  \includegraphics[width=\columnwidth]{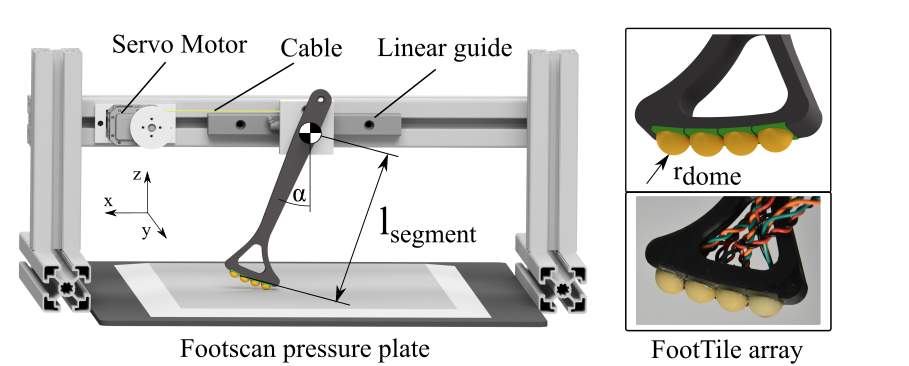}
	\caption{Robotic leg equipped with an array of four FootTile sensors to measure the pressure distribution during locomotion. The foot rolls on a footscan pressure plate as ground truth. The linear guide is pulled by a servo motor to ensure reproducible experiments.}
	\label{fig:sensorArray}
\end{figure}In order to measure the pressure distribution and center of pressure (COP) along a footpad, we implement an array of FootTiles. The center of pressure describes the point where the resultant force vector of a pressure field acts on a body \cite{Cavanagh1978}. To simplify the experiment, we restrict the foot to only roll on the ground in the sagittal plane. Therefore we only use a one-dimensional array of sensors. The foot of our robot \cite{Ruppert2019} is redesigned to have a constant radius of \SI{150}{mm} measured from the ankle joint over the footpad. The footpad arc spans the angle the robot leg sweeps during actuated hopping. We place four sensors along the arc of the foot segment arc. As ground-truth the foot rolls onto a pressure plate (Advanced footscan pressure plate, \textit{rsscan}). The FootTiles are connected to a I$^2$C multiplexer (TCA9548A, \textit{Texas Instruments}), to use several sensors on one I$^2$C bus and reduce the delays caused by the sensor conversion time. To measure the foot position and segment angle, we place two markers along the leg axis. The experiments are recorded with a camera at \SI{50}{fps}. From the video\footnote{\url{https://youtu.be/X7irusBHC3Q}}, we can extract the marker position to calculate angle and position data of the foot segment. The pressure plate data is recorded with \SI{330}{Hz} as well as the FootTile array data. The leg's joint is connected to a linear rail with two ball bearings (\autoref{fig:sensorArray}). The linear rail guarantees that the arc on the footpad of the leg segment experiences pure rolling without height changes. To make the experiment reproducible, we implement a servo motor (MX-28AR \textit{Dynamixel}) that pulls the linear guide sled with a constant velocity by a Teflon cable.

\subsection{Granular and muddy substrate}
\begin{figure}
	\centering
  \includegraphics[width=\columnwidth]{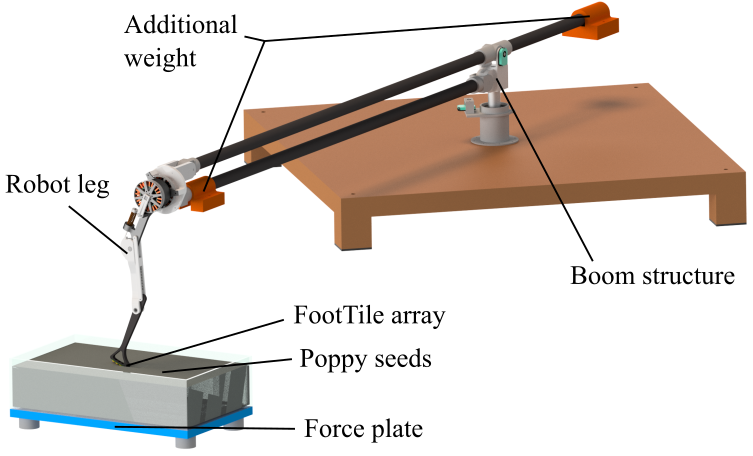}
	\caption{Render of the robot leg with FootTile array. We use poppy seeds as a substitute for a granular medium. The robot is constrained to the sagittal plane by a boom structure. As ground truth for the overall force applied by the leg the poppy seed box is placed on a force plate.}
	\label{fig:boomPoppySeed}
	\vspace{-4mm}
\end{figure}
Here, we compare the ground reaction force (GRF) estimation capabilities of the FootTile array. We simulate locomotion in granular medium by using poppy seeds as substrate \cite{Li2010}. The leg with a total weight of \SI{909}{g} is dropped into a box filled with poppy seeds. As ground-truth we place a force plate (9260AA, \textit{Kistler}) under the box (\autoref{fig:boomPoppySeed}). We record the experiment with a high-speed camera from the sagittal plane at \SI{1000}{Hz}. To simulate locomotion, a servo motor (MX-64, \textit{Dynamixel}) is connected to the hip of the robot to achieve realistic leg swinging behavior. The center of mass (COM) motion of the robot is constrained to the sagittal plane by a boom structure. The leg is dropped from a foot height of \SI{10}{cm} above the substrate. Data from the FootTile array is recorded at a maximum achievable frequency of \SI{330}{Hz}.\\
We further test the sensor under realistic conditions in rough terrain by replacing the poppy seeds with mud from a nearby forest, to showcase the rugged and waterproof design. The leg is dropped into mud while the same data as before is recorded. After use, the sensors are cleaned with water while remaining fully functional. To make the sensor waterproof, only the back of each PCB had to be covered with a layer of waterproof protective urethane resin (Urethane \textit{Cramolin}). The resin protects the PCB from moisture, chemicals, and abrasion. The domes seal the sensors water- and airtight by design.
\section{RESULTS}
\subsection{Sensor validation}
\begin{figure}
	\centering
  \includegraphics[width=\columnwidth]{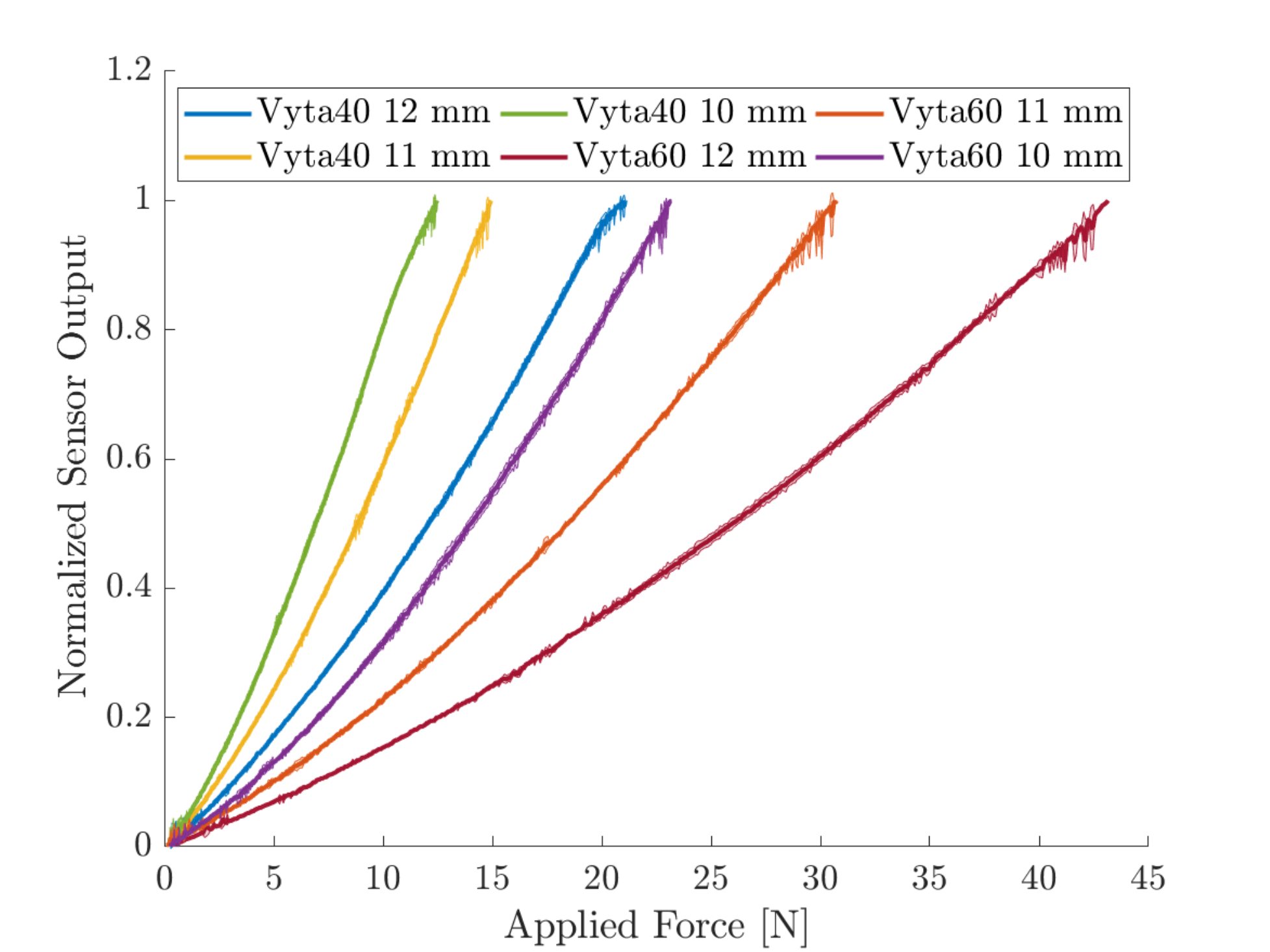}
	\caption{Normalized FootTile pressure data over reference input force  from an external force sensor, for calibration.Selecting stiffer dome material and dome increase the sensor range. With dome material and diameter our FootTile can be adapted for any loadcase. Displayed is averaged data over four experiments with \SI{95}{\%} confidence intervals.}
	\label{fig:forcePressureMolds}
	\vspace{-4mm}
\end{figure}
We compare the data from the FootTile to the force data recorded from the force sensor (\autoref{fig:forcePressureMolds}). The FootTile data is normalized to the maximum value when the barometric sensor saturates. As expected, the FootTiles saturate at different forces depending on the material and dome geometry. The sensors molded with Vytaflex 40 saturate at lower forces than the senors with Vytaflex 60 domes. Sensors saturate at higher forces for bigger dome diameters. All data displayed is averaged over four experiments and displayed with 95\% confidence intervals. As shown in previous studies \cite{Tenzer2014}, the sensor measurements are repeatable with a standard deviation of 0.03 for the normalized sensor output. Because of the deforming bubble geometry, the sensor output is nonlinear with respect to the applied force. The sensor fitting best for the required sensor range of 25-\SI{35}{N} from previous experiments with the robot is the sensor with the \SI{11}{mm} Vytaflex 60 dome. The sensor with PCB and dome has a size of 10 x 10 x \SI{5.5}{mm} and weighs \SI{0.86}{g}. All subsequent experiments use this sensor.\\
Since the measurements have such a small standard deviation, we can approximate the sensor output with a third-order polynomial to correct sensor data in coming experiments. The pressure-force equation for the selected sensor (\autoref{fig:forcePressureMolds} orange) can be approximated with 
\begin{equation}\label{eq:fit}
F(p)=0.13\cdot p^3 +0.02354\cdot p^2+ 0.5702\cdot p + 0.1309
\end{equation}
 with $R_{square}=99.99$\%, where F is the applied force, and p is the FootTile output pressure. 
\subsection{FootTile Array}
\begin{figure}
	\centering
  \includegraphics[width=\columnwidth]{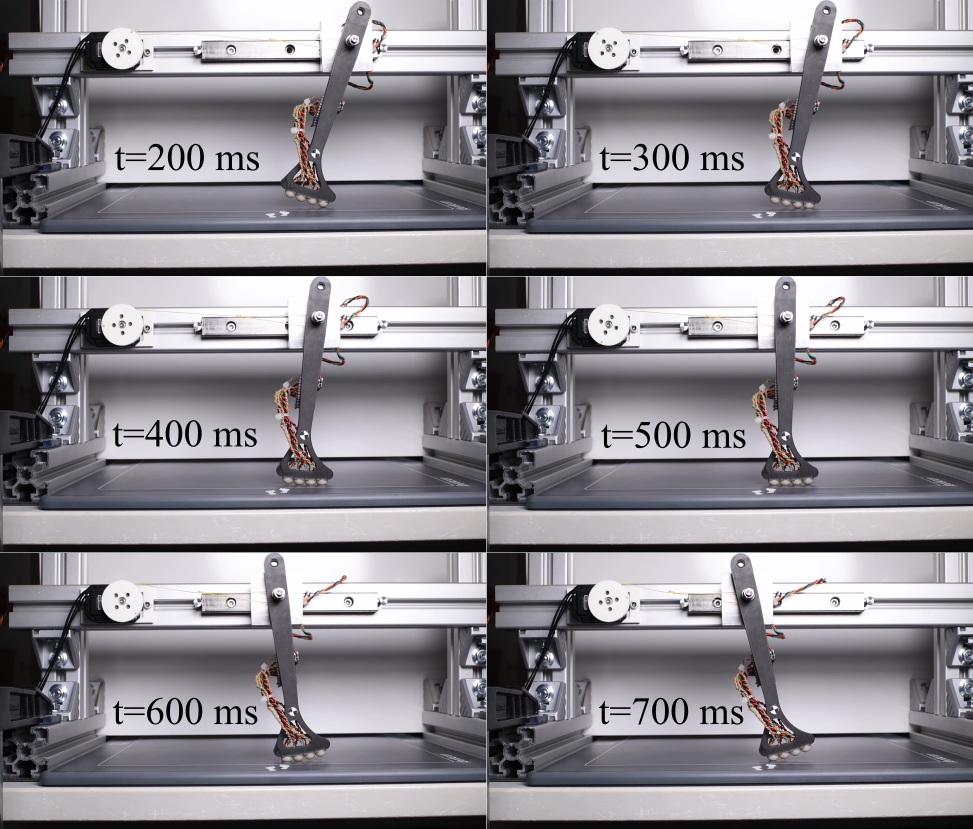}
	\caption{Video snapshots of the FootTile array rolling on the pressure plate for different moments in time.}
	\label{fig:plateExperiment}
\end{figure}
\begin{figure*}
    \centering
	\begin{subfigure}{0.32\textwidth}
	    \centering
	    \includegraphics[width=\textwidth]{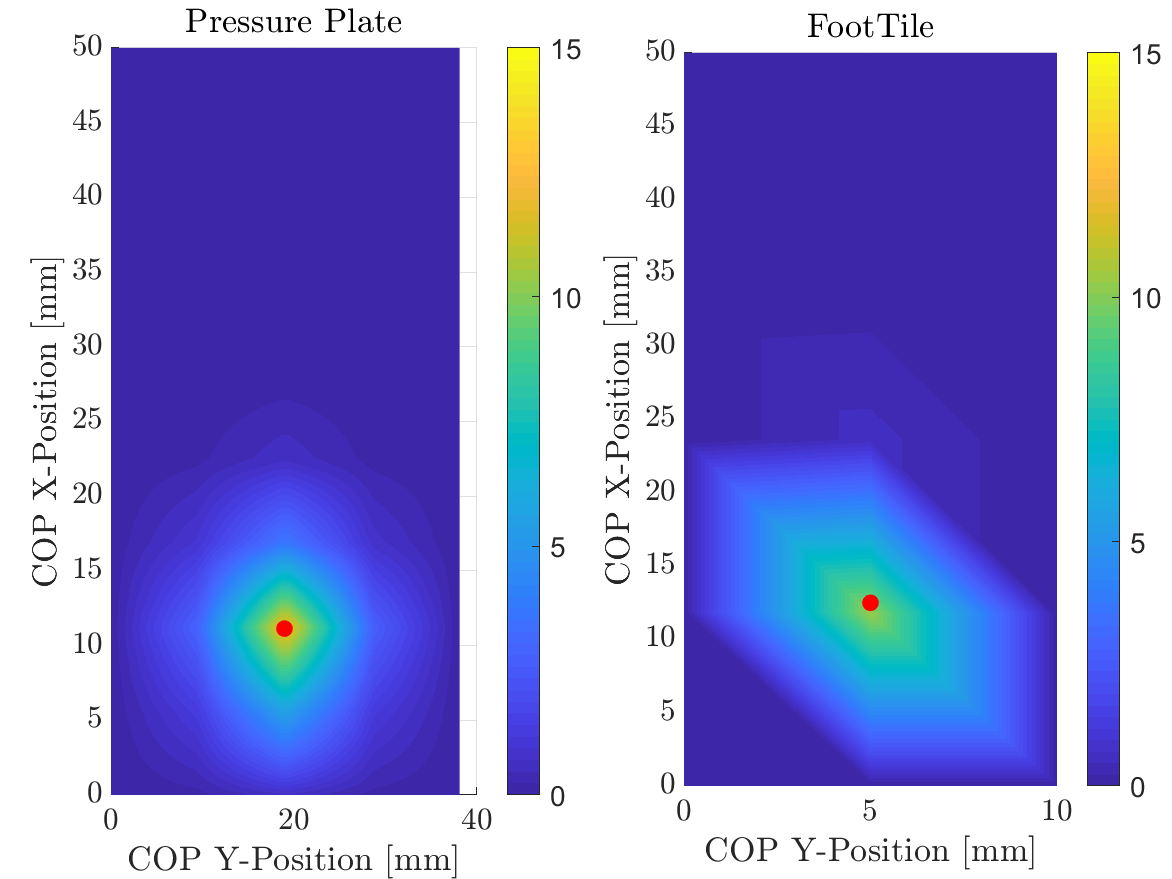}
	    \caption{\SI{200}{ms}}
	\end{subfigure}
\begin{subfigure}{0.32\textwidth}
	    \centering
	    \includegraphics[width=\columnwidth]{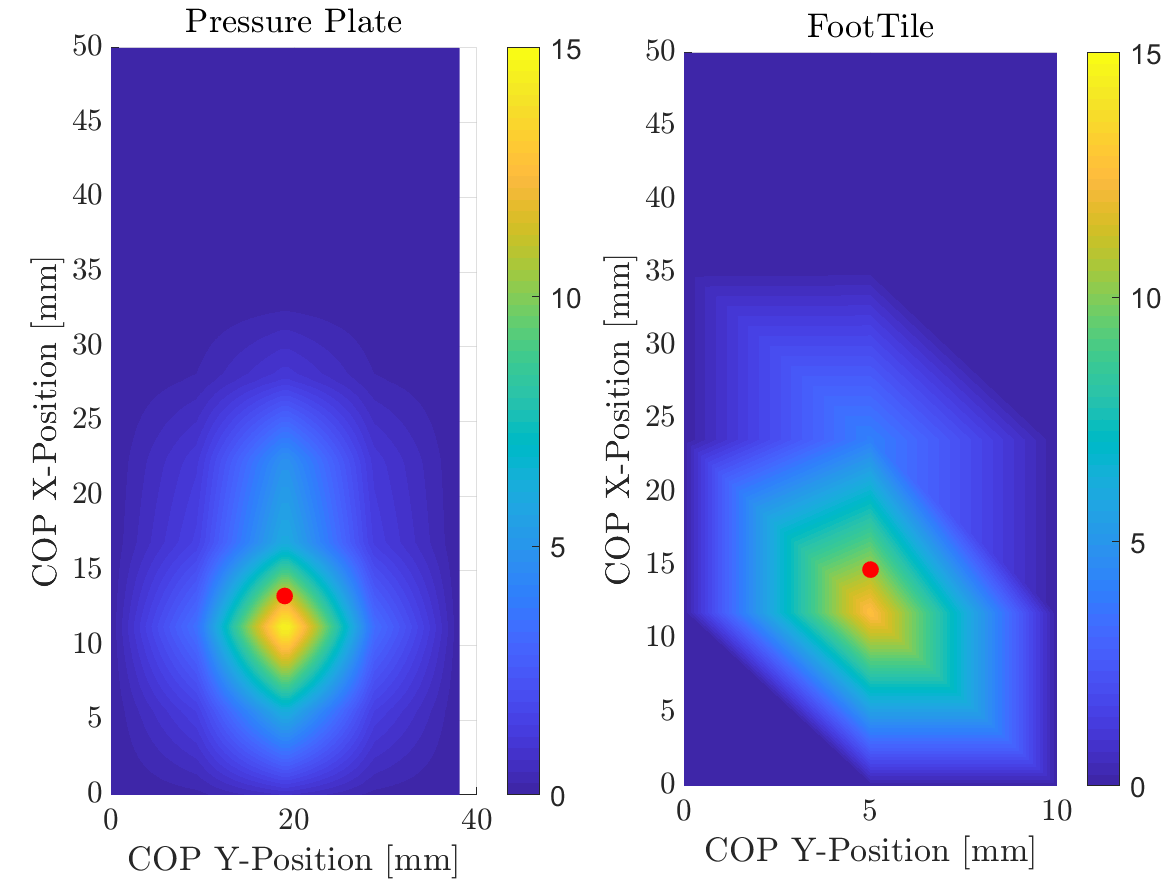}
	    \caption{\SI{300}{ms}}
	\end{subfigure}
\begin{subfigure}{0.32\textwidth}
	    \centering
	    \includegraphics[width=\columnwidth]{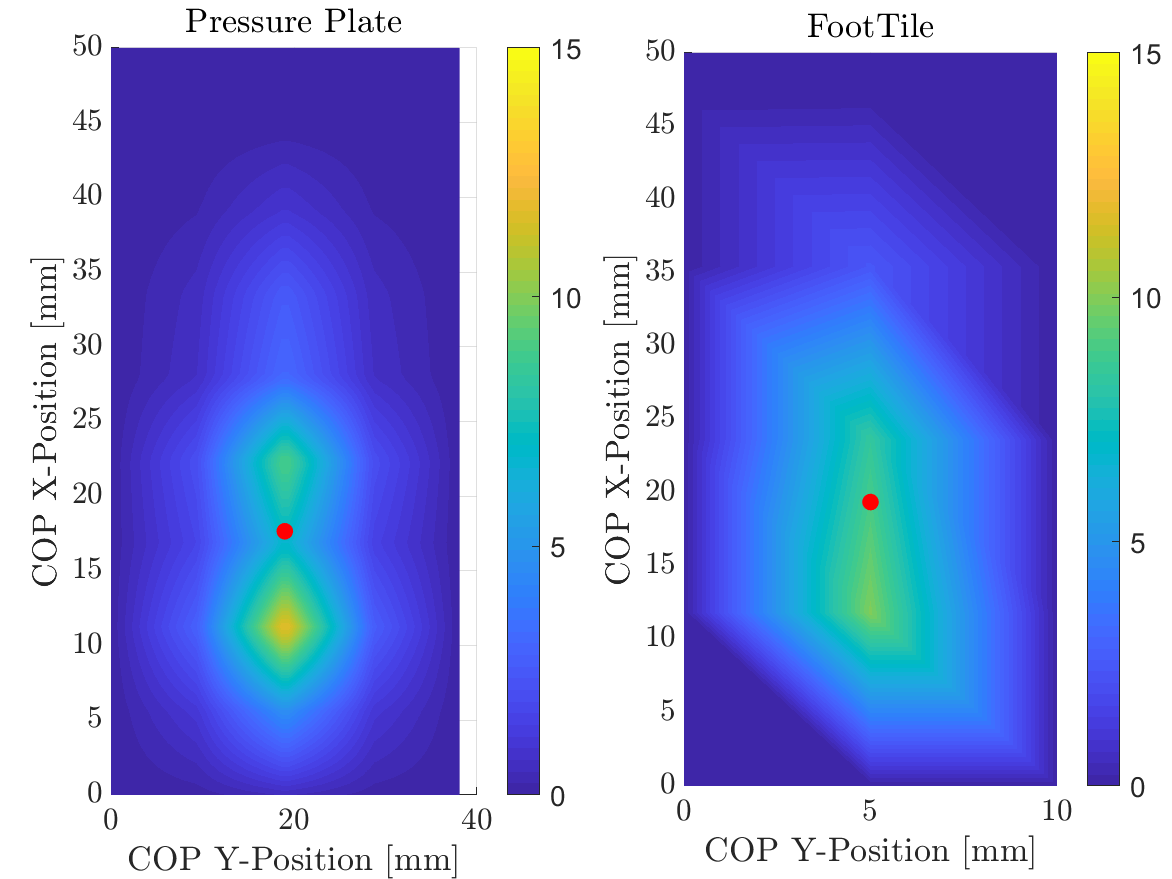}
	    \caption{\SI{400}{ms}}
	\end{subfigure}
\begin{subfigure}{0.32\textwidth}
	    \centering
	    \includegraphics[width=\columnwidth]{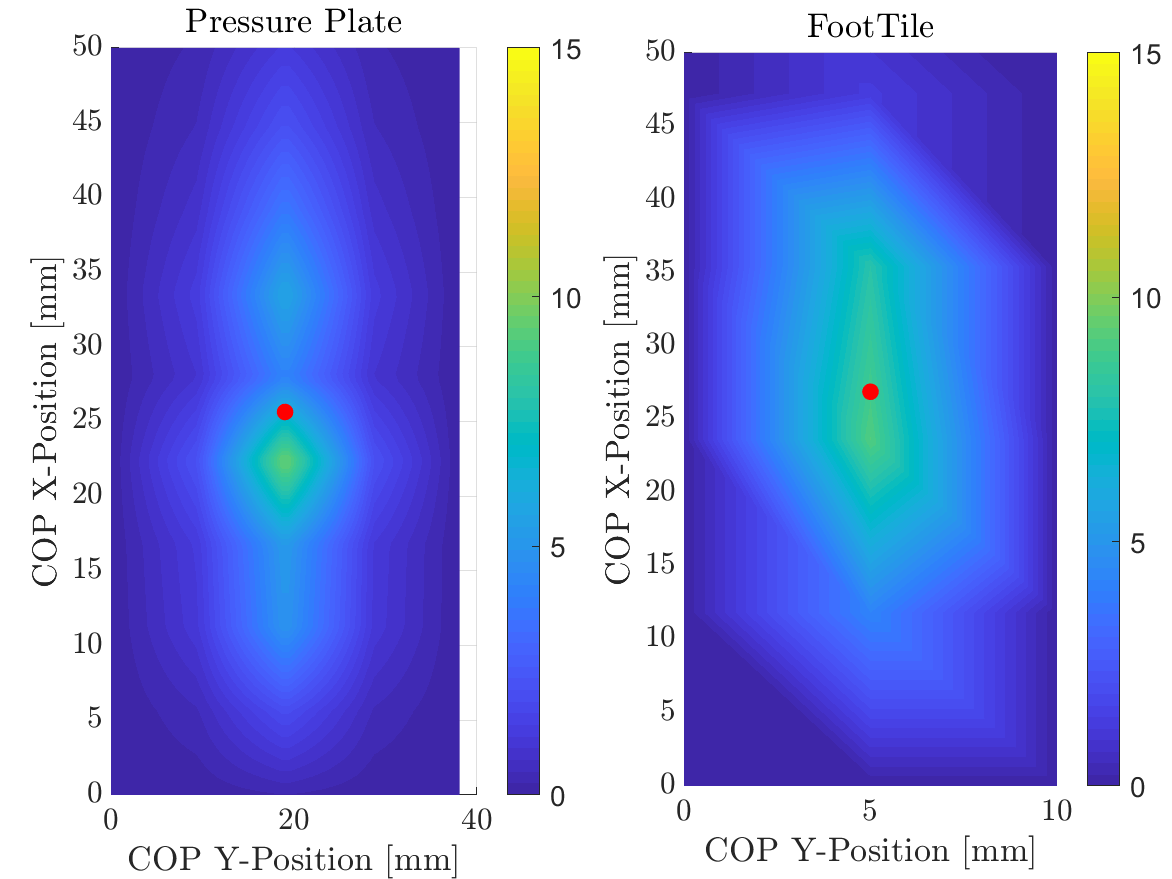}
	    \caption{\SI{500}{ms}}
	\end{subfigure}
\begin{subfigure}{0.32\textwidth}
	    \centering
	    \includegraphics[width=\columnwidth]{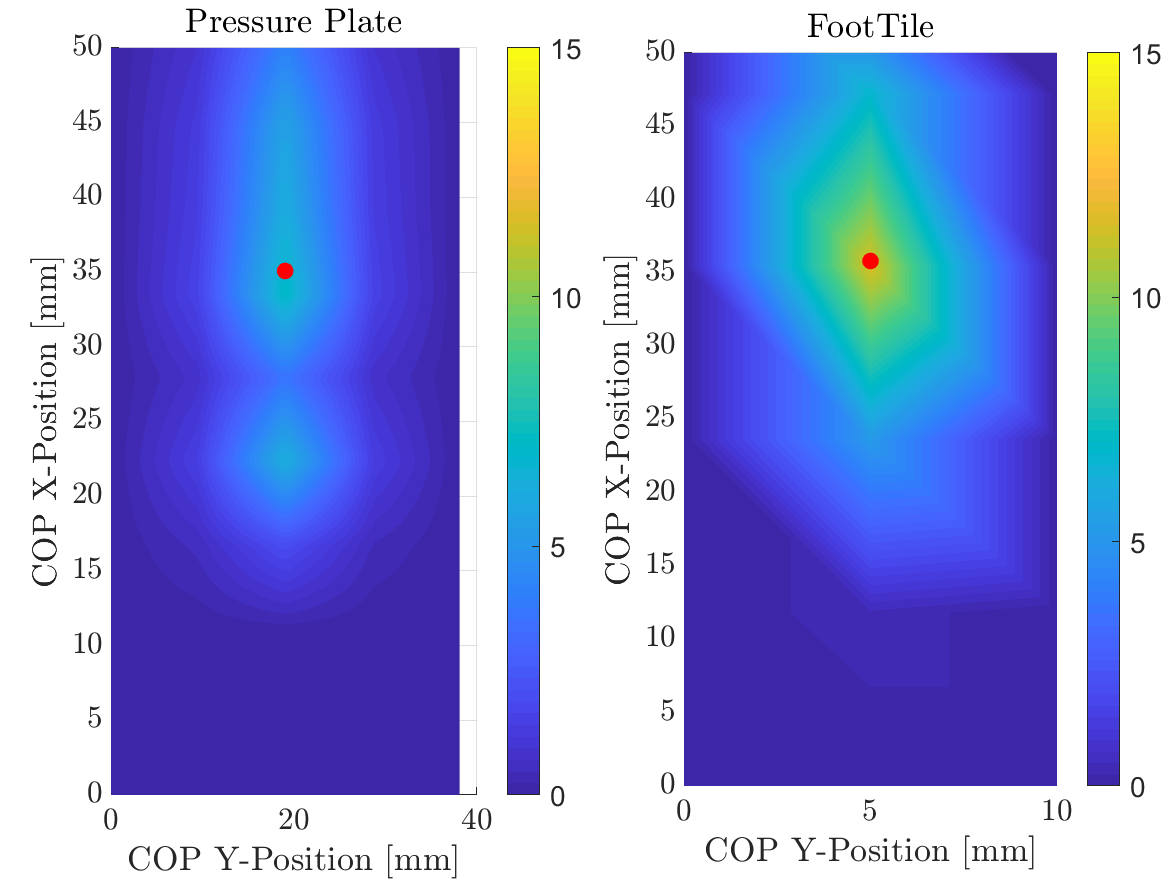}
	    \caption{\SI{600}{ms}}
	\end{subfigure}
\begin{subfigure}{0.32\textwidth}
	    \centering
	    \includegraphics[width=\columnwidth]{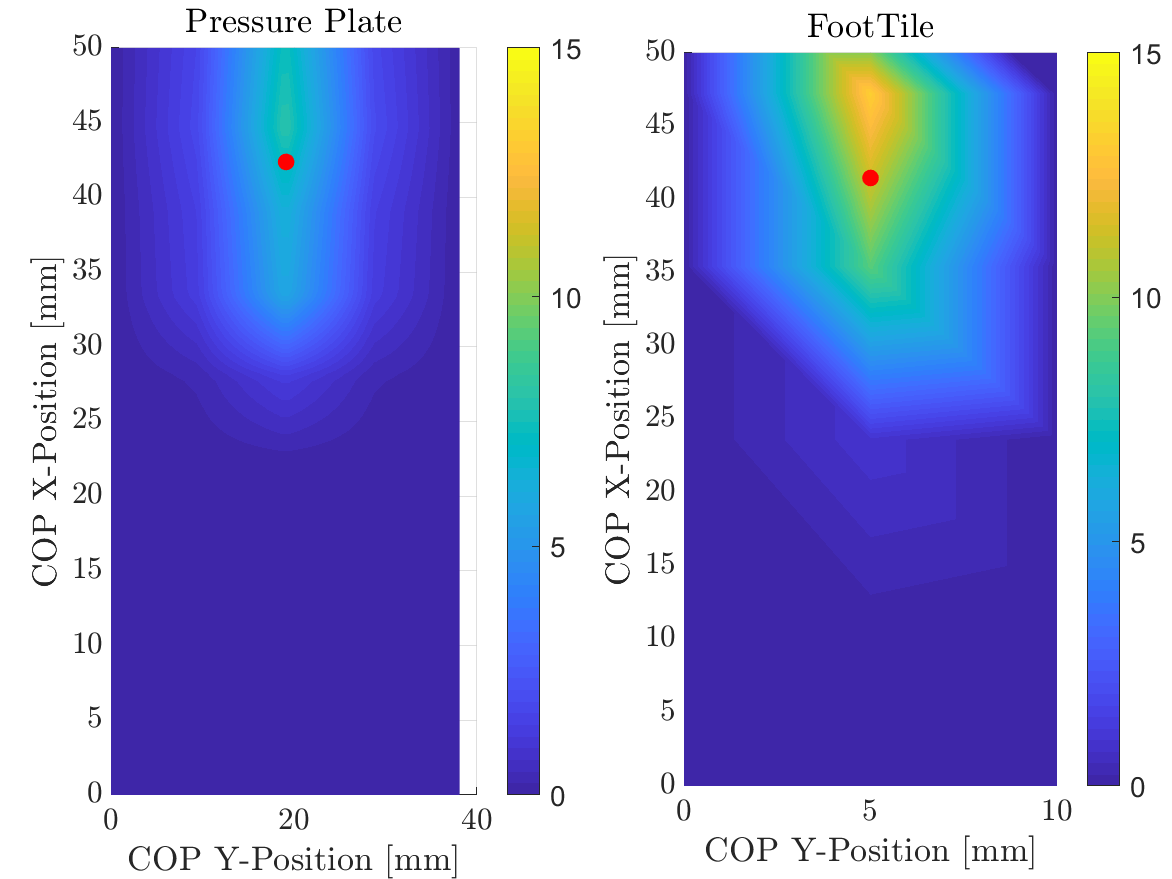}
	    \caption{\SI{700}{ms}}
	\end{subfigure}
\caption{Pressure plate and FootTile array spatial and temporal pressure value readout. Pressure plate data, as ground-truth, shows interpolated data of the active 4x50 sensors. FootTile data shows one-dimensional raw sensor data integrated in a 3x50 grid for comparable visuals. The COP position in mm for each dataset is indicated by the red dot. The error is less than the size of a single FootTile. The small error between ground-truth and FootTile COP is quantified in \autoref{fig:COPyComparison}.}
	\label{fig:COPGrid}
	\vspace{-4mm}
\end{figure*}
\begin{figure}
	\centering
  \includegraphics[width=0.9\columnwidth]{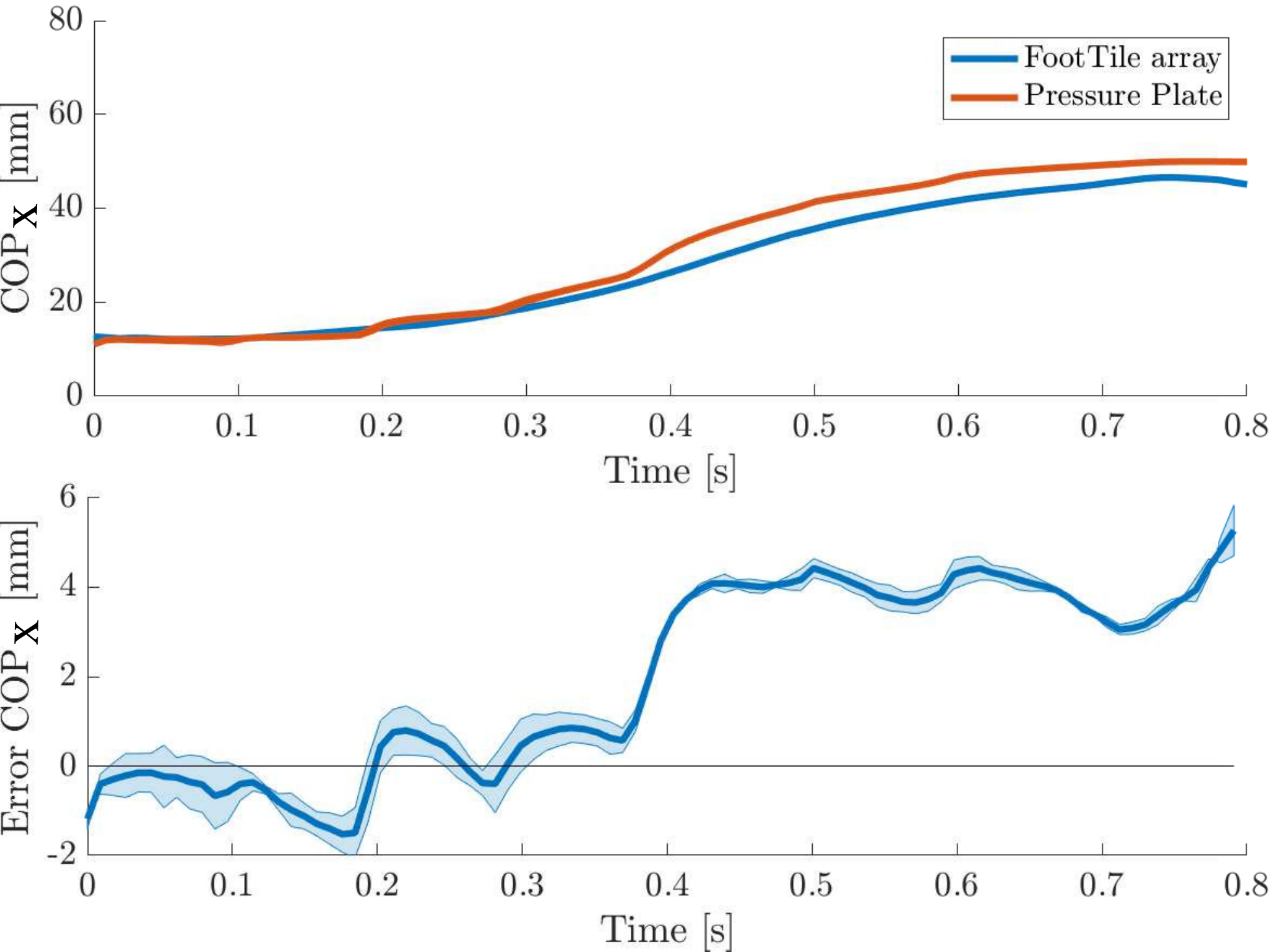}
	\caption{(A) COP x-coordinate for both the FootTile array and the pressure plate as ground truth. The displayed data is the average of four experiments.
	(B) Difference in COP x-coordinate. The displayed data is the average of four experiments with 95\% confidence interval. The error during stance is smaller than \SI{4}{mm}. At the end of the experiment the sensor array loses contact to the pressure plate and the COP estimation becomes less accurate.}
	\label{fig:COPyComparison}
	 \vspace{-4mm}
\end{figure}

To validate how well the FootTiles can estimate the pressure distribution, we plot the pressure plate data versus the data of a $4\times 1$ array of FootTiles (\autoref{fig:COPGrid}). The pressure plate data shows an interpolation of all the active sensor cells. The FootTile data shows the raw data from the individual FootTiles. We calculate the active area of the FootTile array by using the video data (\autoref{fig:plateExperiment}) to calculate the leg segment angle and find the contact points for all sensors with 
\begin{equation}
\begin{aligned}
    y_{contact} = l_{segment} \cdot \sin(\alpha)-r_{dome} \cdot \cos(\alpha) \\
    z_{contact} = -l_{segment} \cdot \cos(\alpha)-r_{dome} \cdot \sin(\alpha) 
\end{aligned}
\end{equation}
where $y_{contact}$ and $z_{contact}$ are the y and z contact point coordinates with respect to the leg segment joint, $l_{segment}$ is the segment length, $\alpha$ is the segment angle to the vertical, and $r_{dome}$ is the dome radius as shown in \autoref{fig:sensorArray}. The origin is in the rotary joint on the linear guide.
To calculate the COP position in the sensor grid, we use a weight function in x and y direction,
\begin{equation}
\begin{aligned}
x_{COP}=\frac{1}{p_{total}}\cdot \sum \limits_{i=1}^m (p(i)\cdot i)\\
y_{COP}=\frac{1}{p_{total}}\cdot \sum \limits_{j=1}^n (p(i)\cdot j)
\end{aligned}
\end{equation}
where $x_{COP}$ and $y_{COP}$ are the Cartesian COP coordinates (\autoref{fig:COPGrid}) in the $m \times n$ grid, $p_{total}$ is the normalized sum of all grid pressure, and index $i$ and $j$ of the grid are used as the linear weight functions in x and y directions. We construct a $3\times 50$ matrix with the raw FootTile data in the second column to be able to compare the pressure distributions visually.
Since the FootTile array in this paper is one-dimensional, we only compare the x-component of the COP position (\autoref{fig:COPyComparison}). The COP estimation from the FootTile array fits the ground truth from the pressure plate. The error is less than $\SI{4}{mm}$ throughout the whole experiment. Only at the end of the experiment, the error gets bigger due to the FootTile lifting from the pressure plate.\\

\subsection{Granular and muddy substrate}
\begin{figure}
    \centering
    \includegraphics[width=\columnwidth]{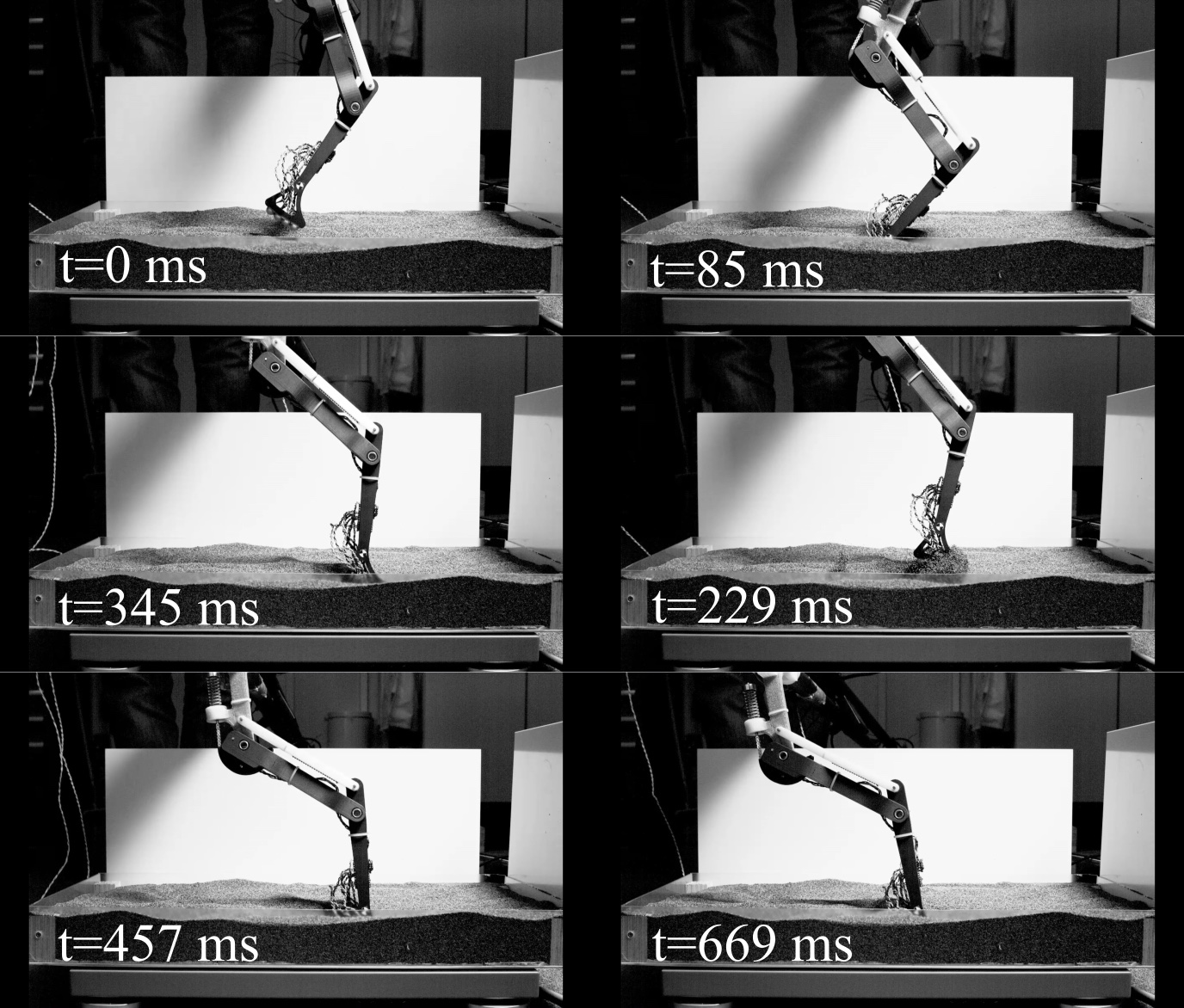}
    \caption{High-speed snapshots of the FootTile leg hopping in the poppy seed box. The leg hops three times with a different number of FootTiles engaging with the ground. Timestamps are displayed in the figure.}
    \label{fig:poppySeedSnapshots}
     \vspace{-4mm}
\end{figure}

\begin{figure}
    \centering
    \includegraphics[width=\columnwidth]{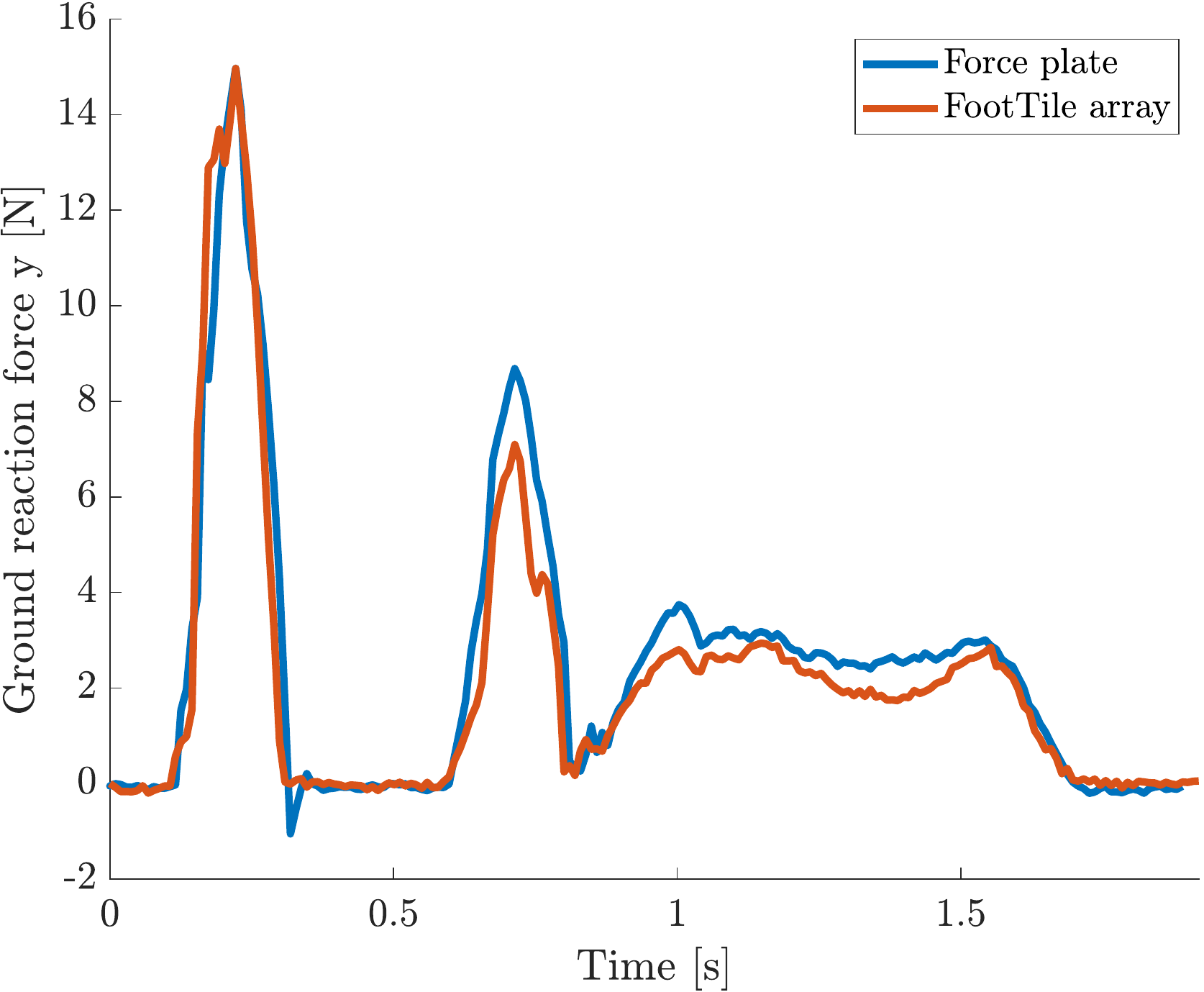}
    \caption{Comparison of the raw force plate vertical ground reaction force as ground truth (blue) and the estimated FootTile forces (orange) from raw pressure data. Mean error is smaller than \SI{1}{N}.}
    \label{fig:GRFcompare}
     \vspace{-4mm}
\end{figure}

We compare the reading of the FootTile array with the force plate recorded ground reaction forces during actuated stepping into the poppy seed box (\autoref{fig:poppySeedSnapshots}). The sensor values are recalculated into forces using \autoref{eq:fit}. The sum of all forces is plotted together with the vertical ground reaction force of the force plate (\autoref{fig:GRFcompare}).\\

\begin{figure}
    \centering
    \includegraphics[width=\columnwidth]{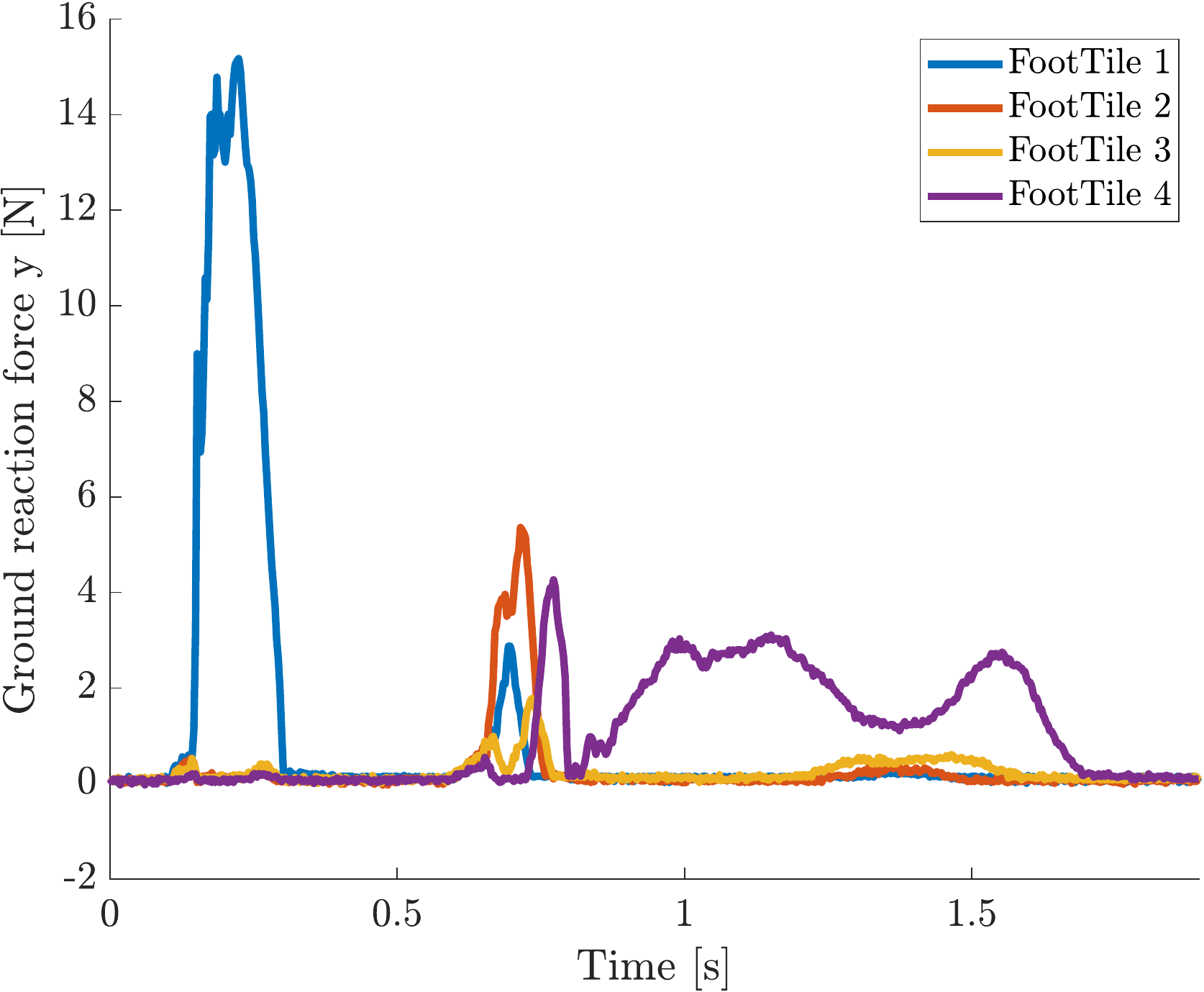}
    \caption{Individual FootTile contribution from \autoref{fig:GRFcompare} to the ground reaction force estimation in granular medium. During the first hop mostly the first sensor is excited. During the second hop all sensors are excited, during the third hop mostly the fourth sensor is excited. }
    \label{fig:GRFcompare2}
     \vspace{-4mm}
\end{figure}
The leg hops three times (visible in supplementary video). During the first hop (\autoref{fig:GRFcompare2}), a single FootTile measures most of the force. During the second hop, the force is distributed among all sensors. During the third hop, the leg comes to rest in the poppy seed box. During all three hops the GRF estimation is reasonable and follows the ground truth data of the force plate. Around the force peaks, there is a small deviation in between the FootTile and the ground truth. We believe this to be caused by the hyper-elastic material properties of the polyurethane dome as well as the torque influences that we neglect here.\\
The FootTile array is able to estimate the vertical ground reaction force with a mean error of less than \SI{1}{N}. Discrepancies between the ground truth and the FootTile estimation stem from the one-dimensional sensor array. Since the sensor is modular, it is possible to use multiple lines of sensors shifted by half the sensor length to increase the sensing resolution in one direction. This way, more sensors are engaged at the same time and potential error during the transition from one sensor to the next could be prevented.\\
\begin{figure}
    \centering
    \includegraphics[width=\columnwidth]{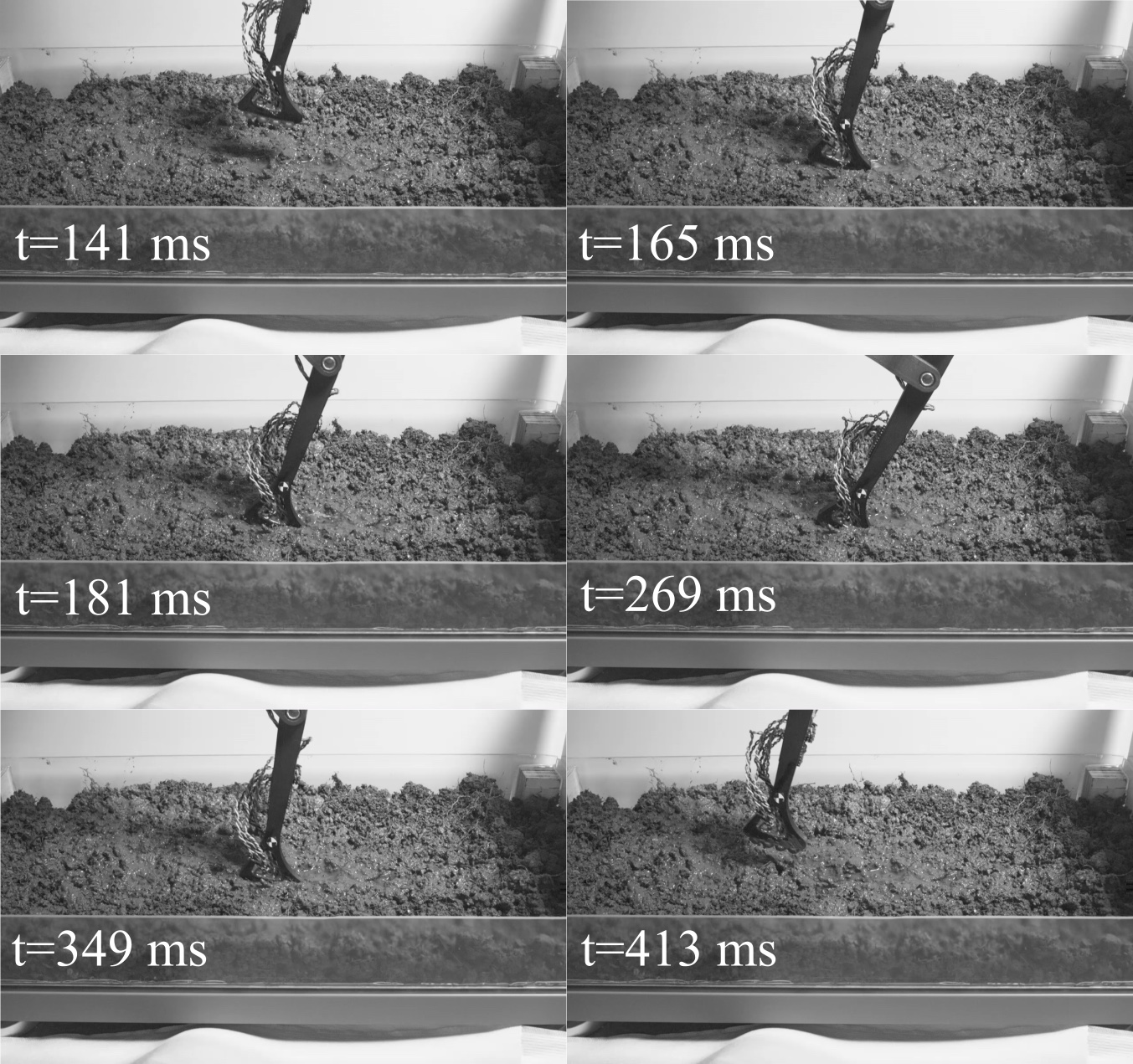}
    \caption{High-speed screenshots of the FootTile array hopping in mud. The waterproof FootTile array is fully submerged in water and mud and remains functional.}
    \label{fig:mudA}
    \vspace{-4mm}
\end{figure}
\begin{figure}
    \centering
    \includegraphics[width=\columnwidth]{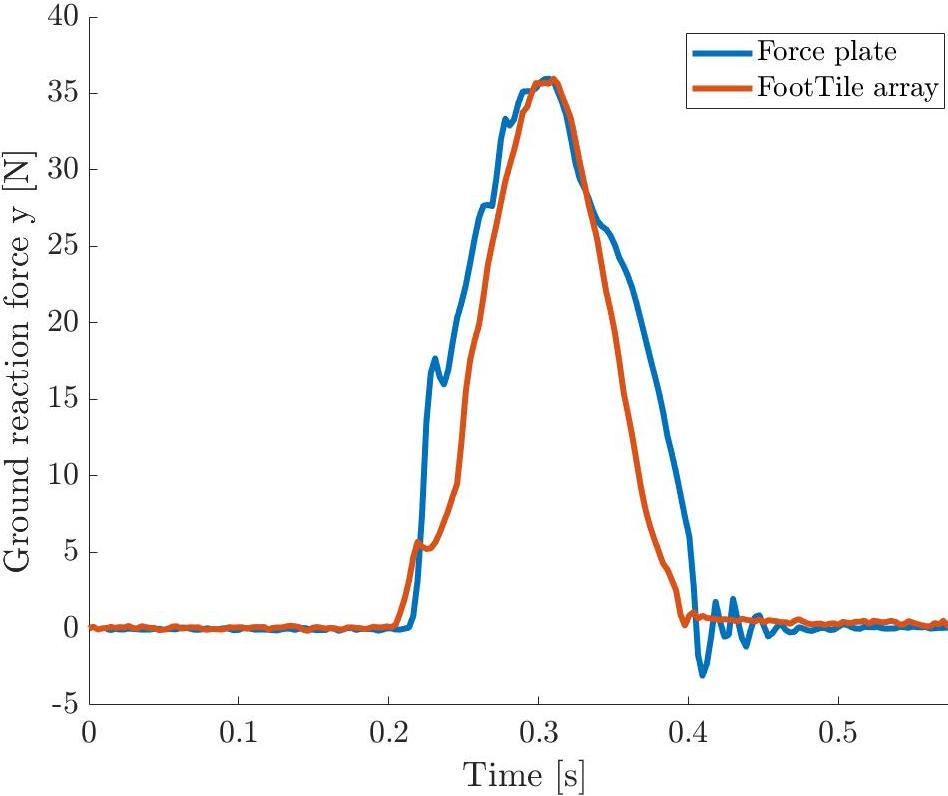}
\caption{(Ground reaction force estimation of the FootTile array hopping in mud.}
    \label{fig:GRFmud}
     \vspace{-4mm}
\end{figure}
To showcase the ruggedness, the FootTile array is dropped into mud, where the sensor array is fully submerged in liquid mud (\autoref{fig:mudA}). The array stays fully functional during the experiment (\autoref{fig:GRFmud}). After use, the FootTile array can be washed in water without compromising the functionality of the sensor (supplementary video). Again, the summation of FootTile GRF estimation follows the ground-truth from the force plate. Small deviations could stem from the highly anisotropic material that includes small stones and roots. This shows the capability of the FootTile to be used outside of a laboratory environment in soft and moist real world terrains.

\section{DISCUSSION}
In this paper, we present FootTile, a force and pressure sensor that is lightweight, small, portable, rugged, modular, and low cost. The sensor can easily be adapted to any required load case. We adapt the previously presented sensor design in size, measuring range, sampling frequency and reduced complexity. We present experimental data and show that FootTile can be used as an alternative to standard biomechanical tools like force plates and pressure plates. FootTile can accurately estimate ground reaction forces as well as center of pressure position in one device.\\
If required, the reading frequency could be improved with a dedicated microcontroller (in the boundaries of material responsiveness). With the current development of smaller MEMS pressure sensors for smartphones the sensor could be miniaturized further. In the future, we plan to mount the small-sized FootTile sensor units on robotic feet as well as animal feet to investigate force and pressure distribution in rough terrain and granular media in the dynamic locomotion of robots and running birds.\\

\section*{ACKNOWLEDGMENT}
We thank Alborz Aghamaleki Sarvestani for the many fruitful discussions in the lab, Annalena Daniels for the material characterization and initial experiments and the Robotics support group (ZWE) for providing the 3D printed parts.
\FloatBarrier
\printbibliography
\end{document}